\documentclass{article}

\usepackage[utf8]{inputenc} 
\usepackage[T1]{fontenc}    
\usepackage{hyperref}       
\usepackage{url}            
\usepackage{booktabs}       
\usepackage{amsfonts}       
\usepackage{nicefrac}       
\usepackage{microtype}      
\usepackage{graphicx}
\usepackage{subfigure}
\usepackage{multirow}
\usepackage[square,sort,comma,numbers]{natbib}
\usepackage{makecell}
\usepackage[final]{nips_2017}

\title{Robust Tracking Using Region Proposal Networks}

\author{
Jimmy Ren \hspace{0.03in} Zhiyang Yu \hspace{0.03in}  Jianbo Liu \hspace{0.03in} Rui Zhang \hspace{0.03in}
\AND
Wenxiu Sun \hspace{0.03in} Jiahao Pang \hspace{0.03in} Xiaohao Chen \hspace{0.03in} Qiong Yan\\\\
SenseTime Group Limited\\
rensijie@sensetime.com
}

\begin{document}

\maketitle

\begin{abstract}
  Recent advances in visual tracking showed that deep Convolutional Neural Networks (CNN) trained for image classification can be strong feature extractors for discriminative trackers. However, due to the drastic difference between image classification and tracking, extra treatments such as model ensemble and feature engineering must be carried out to bridge the two domains. Such procedures are either time consuming or hard to generalize well across datasets. In this paper we discovered that the internal structure of Region Proposal Network (RPN)'s top layer feature can be utilized for robust visual tracking. We showed that such property has to be unleashed by a novel loss function which simultaneously considers classification accuracy and bounding box quality. Without ensemble and any extra treatment on feature maps, our proposed method achieved state-of-the-art results on several large scale benchmarks including OTB50, OTB100 and VOT2016. We will make our code publicly available.
\end{abstract}

\section{Introduction}
\label{Introduction}

Visual tracking is one of the fundamental tasks in computer vision which finds many applications in areas such as video surveillance, robotics, etc. Though the extensive studies on this topic in the past decade, several characteristics in real-life scenarios such as pose variation, occlusion and illumination changes posed considerable challenge on tracker's ability to model object appearance. Many previous tracking methods either used generative models to describe the target appearance or used carefully designed features to build classifiers to distinguish between the foreground and the background. Despite the great success in some settings, the strong assumptions of many statistical models as well as the difficulties in hand-crafting robust features do not make them widely applicable across complex scenes.

The richness of the feature representations learned by deep Convolutional Neural Networks (CNN) is one of the most important reasons underlying its success in many computer vision tasks such as image classification and object detection. This property also movtivated researchers to adopt CNNs as strong feature extractors in the setting of visual tracking. CNN based tracking methods recently achieved outstanding results in terms of tracking accuracy and robustness \cite{nam2016mdnet,nam2016tcnn}. However, the backbone networks used in most of the existing CNN based methods were trained for large scale image classification tasks which are drastically different from visual tracking \cite{Wang2015iccv,Ma2015iccv,Hong2015icml,nam2016mdnet,nam2016tcnn}. Due to such difference, researchers must resort to additional procedures to bridge the gap between the two domains. For instance, ensemble methods \cite{Li2014bmvc,nam2016tcnn} and cross-layer feature selection \cite{Wang2015iccv} are two effective approaches. However, ensemble methods are usually time-consuming because multiple CNNs need to be updated during online training. For cross-layer feature selection, it is also very difficult to manually pick the best lower level feature that works well across different scenes.

In this paper, we aimed at eliminating these extra treatments. We explored the top layer feature maps of a Region Proposal Networks (RPN) \cite{RenSQ16} pre-trained for object detection. We found that due to the connection between proposing interest regions and discriminative tracking, RPN's top layer features are highly relevant and possesses strong discriminative power for tracking. However, the direct utilization failed to generate promising results. By carefully exploring the internal structure of the top layer feature maps, we discovered that such potential can be unleashed by a novel loss function which simultaneously considers classification accuracy and bounding box quality. Without any extra treatment to the feature maps, our proposed method achieved state-of-the-art results in both OTB \cite{Wu15pami} and VOT \cite{vot2016results} benchmarks.

The contributions of our work can be summarized as follows.
\vspace{-0.5\baselineskip}
\begin{itemize}
   \item First, we discovered that based on the top layer feature maps of a RPN we are able to achieve state-of-the-art tracking results without ensemble and using any extra feature engineering.
   \vspace{-0.5\baselineskip}
   \item Second, we showed that in order to achieve such results we need to adopt a novel loss function which simultaneously considers classification accuracy and bounding box quality.
\end{itemize}

\section{Related Work}
Modeling object appearance is the key of many tracking models. To achieve this, generative methods were proposed to maintain the target appearance model \cite{Bao2012cvpr,Jia2012cvpr,Han08pami}. On the other hand, discriminative methods view the problem differently and aim to build classifiers to distinguish between targets and background \cite{Babenko11pami,Hare2011iccv,Gall11pami}. Our proposed method can be categorized as a discriminative method. Previous discriminative methods often used hand-crafted features such as template, histogram features, Haar-like features to build classifiers. Such features are not robust enough to handle large appearance variations such as occlusion and illumination change as well as video quality degradation such as focal blur and low resolution.

One of the fundamental ideas of deep learning is that robust features can be automatically learned from data instead of manually designed by empirical experience \cite{Goodfellow-et-al-2016-Book}. Therefore, the success of deep learning in many computer vision tasks strongly motivated researchers to explore deep neural networks in the setting of visual tracking as well. An early work was done in \cite{Wang2013nips} which performed unsupervised feature learning on images to learn general feature representations. The advantage of this method is only unlabeled data is required. Many of the recent studies \cite{Wang2015iccv,Ma2015iccv,Hong2015icml,nam2016tcnn} obtained encouraging results by building trackers directly upon CNNs pre-trained on large scale image classification tasks \cite{Krizhevsky2012nips}, rather than training the deep network from scratch. This progress showed that the rich feature learned from image classification can be useful for tracking.

However, image classification and tracking are two tasks drastically differ in objectives. Because feature extractors in CNNs are shaped by objective function's error derivatives, image classification CNNs do not guarantee to provide relevant features for tracking. Therefore, additional procedures need to be carried out to bridge this gap. In \cite{nam2016tcnn} and \cite{Li2014bmvc}, the authors used an ensemble of CNN models to make the resulting model generalize well. Despite the effectiveness, ensemble methods are usually computationally heavy because multiple CNNs need to be updated during online tracking. Another useful strategy is to use cross-layer feature selection and aggregation \cite{Wang2015iccv}. The issue with this method however is that it is very hard to hand-pick the best group of lower level feature maps which consistently provides relevant features across scenes and domains.

Multi-domain learning was used in \cite{nam2016mdnet} and achieved very good results. However, unlike our method this method not only requires lots of labeled tracking data which is scarce in real applications, it also used separate models for different datasets. \cite{Zhu2016cvprw} is a recent paper which is closely related to ours. The authors also used a RPN to build trackers. However, they used a much bigger VGG network. The loss function used is also very different. Further, they did not report results on large scale benchmarks such as OTB and VOT which covers many different realistic scenes.

\section{Analysis and Our Method}

\subsection{Why classification CNN is problematic in tracking}
Because the resulting feature detectors in CNNs are shaped by their error derivatives through back-propagation during training, thus the two networks would possess very different feature detectors if their error derivatives behaves differently \cite{Goodfellow-et-al-2016-Book}. A random guess of a 1000-way classification model will have much larger chances to misclassify a certain class label than a 2-way classification model during training. Thus the former's error derivatives are a lot bigger. At the same time, the error derivatives of the 1000-way classification model is also much noisier because the samples in a mini-batch usually only cover a portion of all the class labels. Though large classification CNN is able to provide rich features, the process of shaping these features are almost completely different from what discriminative tracker really cares about, namely the ability to distinguish between the two particular class labels, the foreground and the background. Therefore we hypothesized that without extra treatments on the feature maps, large classification CNNs are not able to achieve good results in tracking.

\paragraph{Tracking using classification CNN without ensemble and feature engineering}
In order to test our hypothesis, we adopted MDNet's \cite{nam2016mdnet} online tracking framework because it does not add any ensemble and feature engineering logic to the CNN and concise enough to plug-in any other existing CNN models to perform fair comparison. We will not cover the details of MDNet due to the limitation of the length of the paper.

We tested this hypothesis using four off-the-shelf CNNs trained for ImageNet, namely AlexNet \cite{Krizhevsky2012nips}, ZF net \cite{Zeiler2014eccv}, VGG-16 and VGG-19 \cite{Simonyan2015iclr} network. Except replacing the backbone network, we kept all the other settings the same with our method. The evaluation was carried out in both OTB and VOT dataset. We found that comparing to the state-of-the-art results these models showed significant performance degeneration. We observed that the drifting problem is very common among videos. When tracking is successful, the quality of the bounding boxes is lower than expected. Another surprising result is that comparing to the much smaller AlexNet and ZF net, VGG net did not bring better performance. In some videos, it performed even worse. This showed that the richer features learned by the bigger CNNs in the classification task may not relevant to the tracking task.

On the other hand, proposing interest regions is conceptually very similar to the goal of discriminative tracking. An ideal interest region is the one contains an object of any kind, so RPN must have the knowledge on both ``objectness'' and background. This is also the key for discriminative trackers. In other words, discriminative tracking can be thought of as an interest region tracking process. The unsatisfactory results obtained by the classification CNNs strongly motivated us to explore RPNs for tracking.

\subsection{Exploring RPN's Top Layer Features}
The success of MDNet is inspiring because it showed that it is possible for a single medium size CNN to achieve outstanding tracking results without feature engineering. MDNet realized this possibility by shaping the CNN's feature detectors using large scale labeled video frames so that the network is able to extract features of high relevance to tracking. However, collecting such labeled data at scale is intractable in practice. Our goal is to eliminate ensemble and feature engineering without the support of any labeled video frames. The basic idea underlying our method is if we are able to use a novel loss function to exploit the conceptual similarity between RPN and tracking, the network should be able to generate relevant features for tracking. We followed the same online tracking framework mentioned in the previous section throughout the this study to explore this idea.

\paragraph{Receptive Field and Input Size}
The first design choice we need to make is to select the proper input image size. The input image size in MDNet is 107x107, this number is fully empirical. We adopt the principle of receptive field \cite{lecun98} to guide our choice. In the context of CNN, the receptive field of a particular neuron refers to the number of related pixels in the input image. For instance, if a neuron is directly generated by a 5x5 convolution filter on the input image, the receptive field of this neuron has 5x5 pixels. Because the input images in our context are large image patches covers the whole object, if the receptive field of the neurons in the selected layer smaller than the input image, the features may be too local and can not capture the object appearance robustly. On the other hand, if the receptive field of the neurons can potentially cover a much bigger image than the current input image, the feature might be too redundant.

In our case, we used a ZF style RPN \cite{RenSQ16} as the candidate backbone network. In this network the $conv5$ layer was trained to generate region proposals. The receptive field of this layer is 171x171. According to \cite{nam2016mdnet}, it is beneficial to let the input image cover a small piece of background content additional to the object of interest, the proper size of the input image is thus 203x203 because in the RPN we used, the stride between two adjacent anchors corresponds to 16 pixels in the input image.

\paragraph{Matching Anchor Boxes for Tracking}
An important contribution of RPN is the design of anchors. An anchor is essentially a 1x1 position in feature maps of a specified layer. For instance, every 1x1 position of the $conv5$ layer in the ZF network. In RPN, the prediction of objectness as well as the bounding regression are performed based on this feature map. Each anchor is conceptually corresponding to a few ``imaginary'' anchor boxes in the original input image. The insight is the scale and the aspect ratio, even the exact position of each anchor box is largely defined by design decisions and only loosely guided by the theory. For instance, we can assign 9 anchor boxes in the original image for each anchor in the feature map to cover the 3 arbitrary types of aspect ratio and 3 arbitrary scales.

\begin{figure}
  \includegraphics[width=0.6\linewidth]{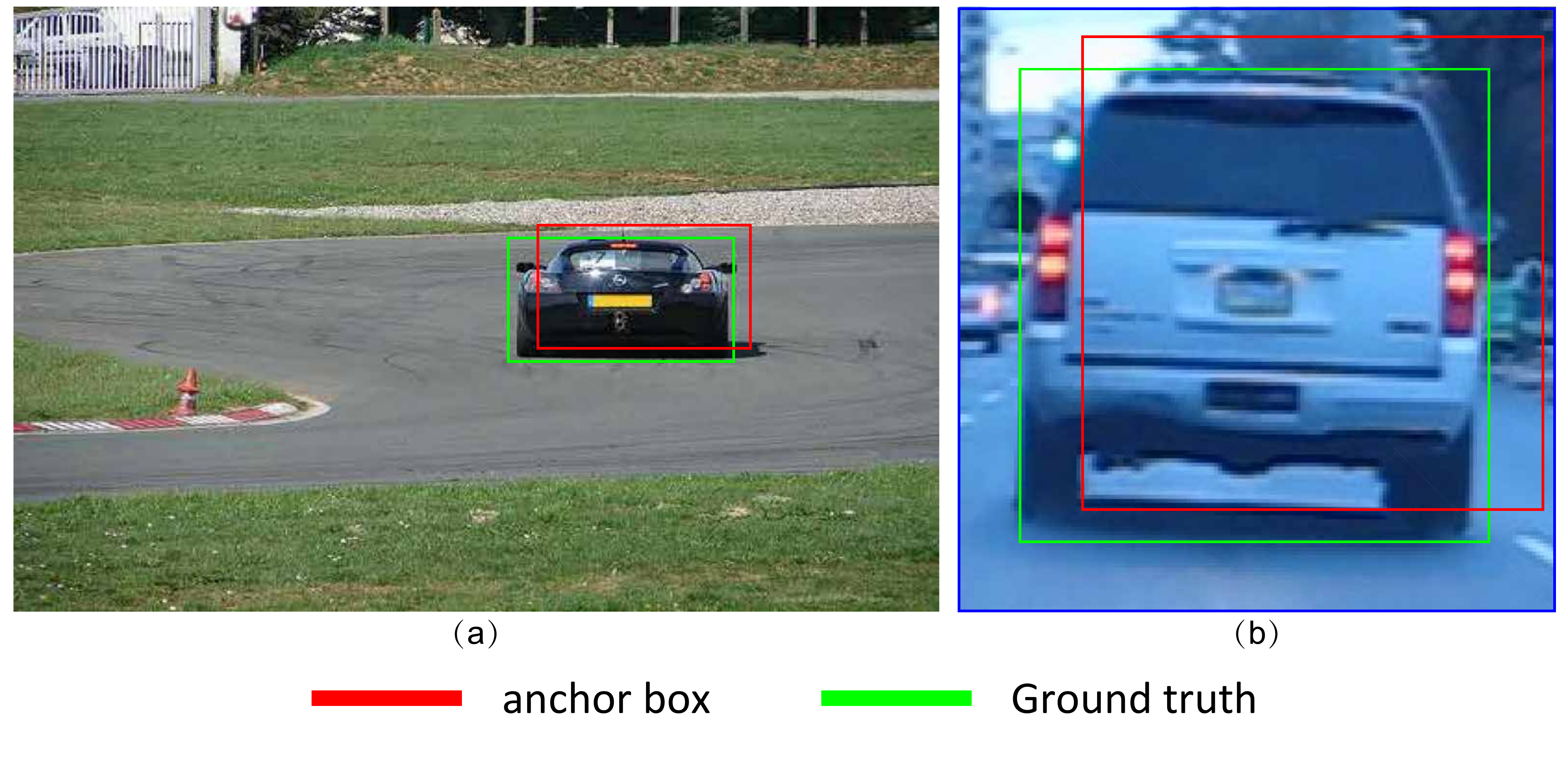}
  \centering
  \caption{Left: The illustration of the matched anchor box and the groundtruth box in the detection setting; Right: The matched anchor box and the groundtruth box in the tracking setting.}
  \label{fig:anchor}
\end{figure}

The purpose of anchor is to provide a foundation based on which the Intersection over Union (IoU) between the groundtruth bounding boxes and the anchor boxes can be computed. This concrete IoU value can therefore be used to categorize each anchor box as possitive box (the object) and negative box (the background). The left diagram of fig.\ref{fig:anchor} shows the groundtruth bounding box and a matched anchor box where the IoU between the two is larger than 0.7. By doing this, each position (anchor) in the $conv5$ layer is endowed with a certain physical meaning, namely whether this anchor should be classified as an object or background. This makes the ensuing classification and bounding box regression meaningful. Please refer to \cite{RenSQ16} for more technical details on anchors.

Because the design of anchor has its flexibility as long as it follows basic design principles, it gives us a chance to adjust the anchor configuration for tracking without modifying the feature detectors in the original RPN. This is critical for our goal because we have to solely rely on the given bounding box in the first frame for model adaptation. Training or fine-tuning a large number of weights is not affordable. Therefore, we adopted the following additional design principles for our anchor design in the tracking task. First, the design should be as simple as possible so that we shall have limited amount of trainable weights to adapt during online tracking. Second, the new anchor configuration should be compatible with the pre-trained feature detectors in RPN so that the network shall extract relevant features for tracking without additional learning.

The right diagram in fig.\ref{fig:anchor} shows the relationship between the groundtruth box and one possible anchor box in our tracking setting. Similarly, if an anchor box is matched with the groundtruth, it should be classified to an object. Following \cite{nam2016mdnet}, it differs from the detection setting that the input images sampled from the scene are always of a square shape in which the groundtruth box is also a square. When the object is bounded by a rectangle, the image will be reshaped to a square so that the same CNN can process different kinds of objects in a unified manner. To utilize this fact, we dropped all the anchor boxes for the aspect ratio of 1:2 and 2:1. This design choice follows the first aforementioned design principle because it significantly reduces the number of matched anchor boxes, thus also reduces the corresponding trainable weights. We also set the size of the anchor box 171x171 which is the same as the receptive field of the neurons in the $conv5$ layer, thus follows the second design principle as well.

\begin{figure}
  \includegraphics[width=0.5\linewidth]{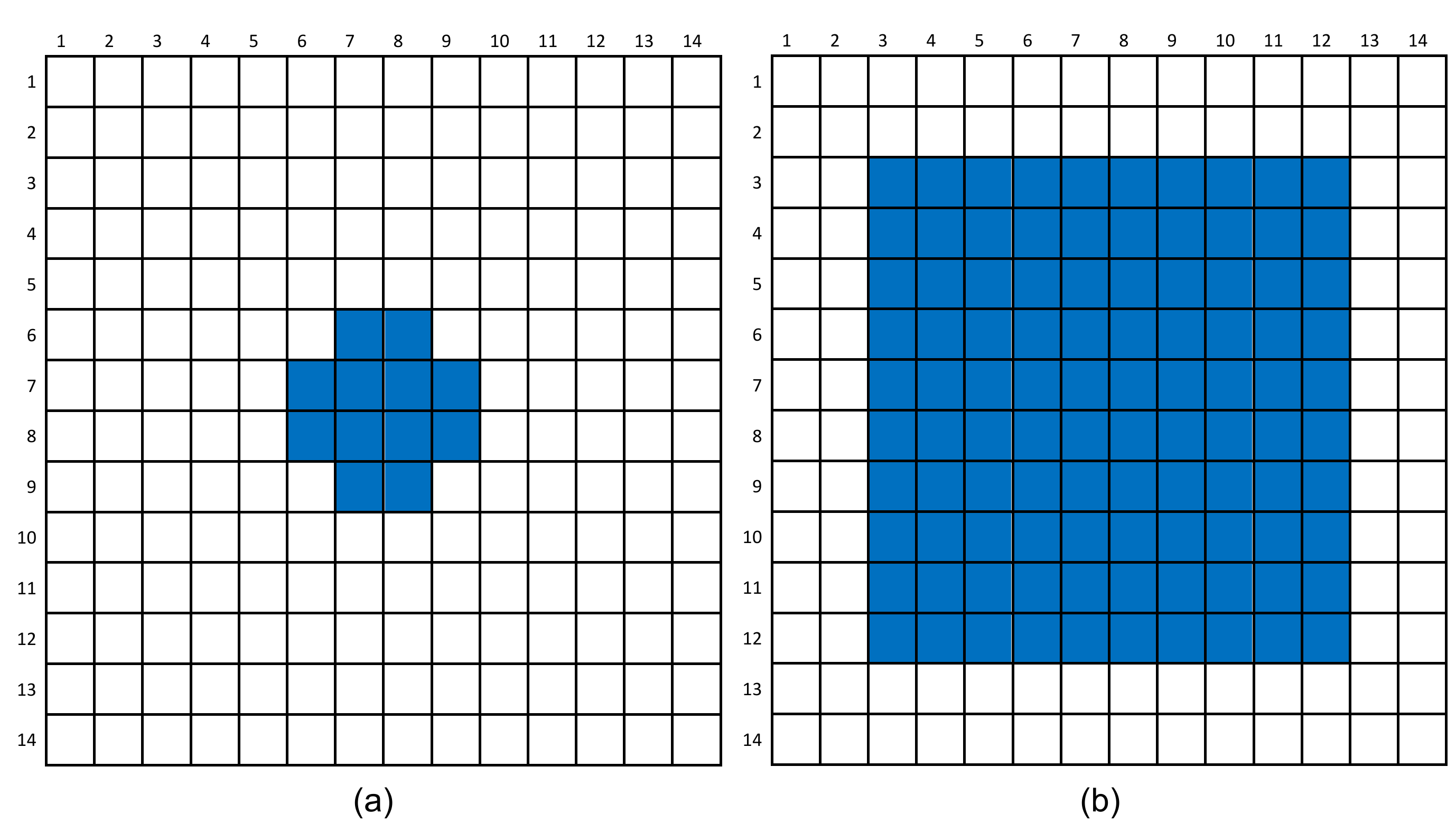}
  \centering
  \caption{Left: View the structure of the top layer feature map as anchors. It shows all the anchors with matched anchor box if the IoU is 0.7; Right: A different perspective. does not assume any structure in the top layer feature map.}
  \label{fig:anchor_choice}
\end{figure}

If the size of the input image is 203x203, the size of the $conv5$ layer in the ZF network is 14x14. Because a 14x14 map does not have a central point, we treat the central 2x2 position the same. It means their corresponding anchor boxes have the IoU of 1.0 with the groundtruth box. If we set the IoU threshold to 0.7, fig.\ref{fig:anchor_choice} (a) shows all the anchors which have the matched anchor box. 

\paragraph{How Tracking Is Done}
The tracking is done as follows. The loss function of RPN can be defined 

\vspace{-0.5\baselineskip}
\begin{equation}
L = \sum\limits_{i} L_{cls}(p_i,p_i^*) + \lambda\sum\limits_{i}p_i^*L_{reg}(t_i,t_i^*),
\label{eq:rpn_loss}
\end{equation}

\noindent where $i$ is the index of an anchor, $p_i$ is the probability prediction of anchor $i$ being and object. $p_i^*$ is the groundtruth label. It is set to 1 if its corresponding anchor box has a match and set to 0 for negative samples. $L_{cls}$ is a classification loss to discriminate the two. We used cross entropy loss in this study, other classification loss should provide similar results. The second term in the equation is a regression loss for bounding box regression. This is only performed for positive samples.

Once the matching is done, the classification term in eq. (\ref{eq:rpn_loss}) can be used to perform online tracking. A simple weighted sum over all the scores is used to get the final score.

Using anchors has two major advantages. First, it provides a vehicle to explore the internal structure of the top layer features. To optimize for tracking performance, different anchor box matching strategies can be tested. Second, each of the matched anchor box can be thought of a data sample because they all contribute to the loss function. Compare to the a single classification output, this approach enlarges the number of training data in a very efficient way. It significantly reduces the risk of overfitting for online training.

\paragraph{Matching Strategy and Tracking Performance}
The labels for positive samples are set to 1 in the matched anchor positions. For the rest of the places in the 14x14 area, they are ignored in computing the loss. The label configuration for negative samples is the same except all the 1s are replaced by 0s.

\begin{figure}
  \includegraphics[width=0.7\linewidth]{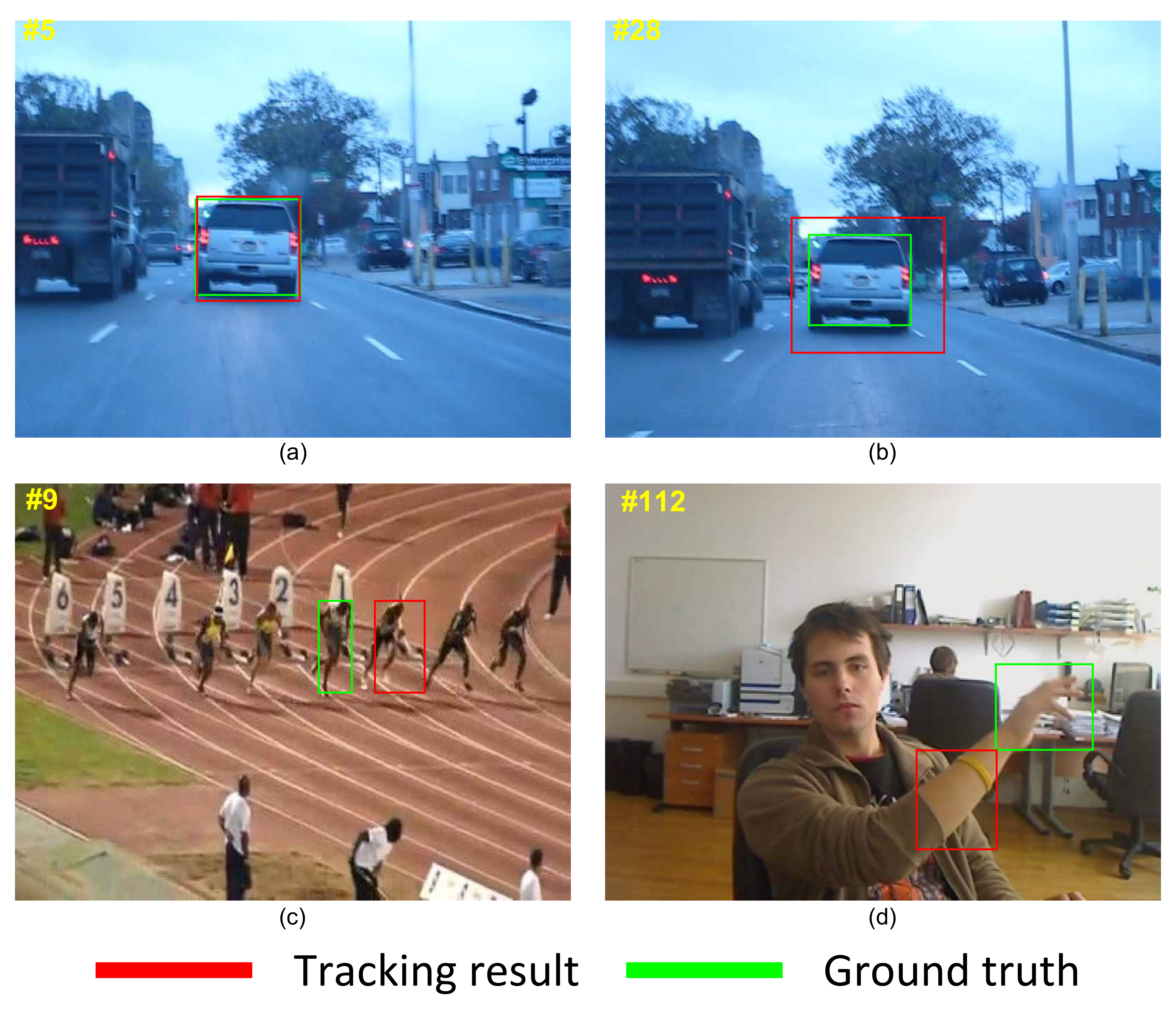}
  \centering
  \caption{The tracking result with the model which does not use the loss function simultaneously considers tracking accuracy and bounding box quality.}
  \label{fig:explore}
\end{figure}

We explored different IoU values from 0.5 to 0.9 for anchor box matching and ran the tracker for experiments in both VOT and OTB benchmark. However, the results were disappointing. We observed the following phenomenons and illustrated them in fig.\ref{fig:explore}. If we use a relatively big IoU threshold in anchor box matching, less anchor positions are used in computing the loss and the loss focused more on the center of the feature maps. We found that the system is able to successfully track most of the objects, including the very challenging ones, at the beginning but the bounding box will gradually get large. Therefore the final score is not high. This phenomenon is showed in fig.\ref{fig:explore} (b). On the other hand, if we use a relatively small IoU threshold, we found improved bounding box quality. However, this only happens for those objects whose appearance does not change dramatically throughout the video (e.g. the car in fig.\ref{fig:explore} (a)). For other objects, such as human body and hand, the tracking drifted from the groudtruth bounding box quickly. This phenomenon is showed in fig.\ref{fig:explore} (c) and (d).

This result is interesting, here are a few insights. The purpose of a big IoU threshold is essentially to view tracking in the region proposal perspective. However, the RPN's original bounding box regression loss seems not to be effective here mainly because we only have labeled bounding box data in the first frame. This makes the bounding box regression loss in eq. (\ref{eq:rpn_loss}) very noisy and prone to making errors. Therefore, the offset in bounding box is expected. Such error is accumulated over frames and caused the phenomenon in fig.\ref{fig:explore} (b). On the other hand, if we disable the bounding box regression term in eq. (\ref{eq:rpn_loss}) and use a low IoU threshold for the classification loss, It is closer to the previous formulation in MDNet. In the extreme case, if all the positions in the 14x14 area are set by a label, it essentially degenerates to the original MDNet setting. While we shall face the inevitable drifting problem in this setting, it actually gives a practical way to safely eliminate the bounding box regression term in eq. (\ref{eq:rpn_loss}) and still generate high quality bounding boxes. The question is how we should leverage both perspectives and leverage the advantages of them respectively.

\paragraph{The New Loss Function}
According to the previous discussion, the loss function we would like to realize should be able to simultaneously consider the tracking accuracy and the bounding box regression. Because if we do not optimize for such an objective, the accumulative error along the process will inevitably lead to tracking failure.

The key idea in this paper is to use a customized version of anchor design to harness the power of RPN so that we can leverage the strong relevance between objectness and tracking to provide a strong foundation for tracking accuracy. Meanwhile, the bounding box quality should also be regularized by a new term which does not suffer from the insufficient labeled bounding box data. The proposed loss function is

\vspace{-0.5\baselineskip}
\begin{equation}
L_{RPN2T} = \alpha\sum\limits_{i} L_{cls}(a_i,a_i^*) + \beta\sum\limits_{i} L_{cls}(q_i,q_i^*),
\label{eq:track_loss}
\end{equation}

\noindent where $a_i$ is the predicted probability of an anchor $i$ being an object under the selected anchor matching scheme. $a_i^*$ is the groundtruth label. $p_i$ and $p_i^*$ are also the predicted probability and the corresponding groundtruth but follows a setup emphasizes the bounding box quality. $\alpha$ and $\beta$ are two variables to balance the two terms.

After a systematic exploration of the matching strategy, the design of $a_i^*$ and $p_i^*$ are illustrated in fig.\ref{fig:anchor_choice} (a) and fig.\ref{fig:anchor_choice} (b) respectively. The value of $\alpha$ and $\beta$ are set to 1 and 10 respectively. We used the same setting for all the videos in the experiments. We will show that this setting generalizes very well across two large scale tracking benchmarks including OTB and VOT.

\subsection{Acceleration by Knowledge Distillation}
One issue with the current design is its speed. The reason why the current network is slow is mainly attribute to the high resolution of the input image. However, according to our analysis, such resolution is necessary for our method to work because it guarantees the resolution of the top layer features. We tried to directly lower the size of the input image to half of its original width and height. Though the whole framework still works to some extend, we also encountered considerable performance degradation. Therefore, we have to come up with a new way to accelerate the tracking procedure without losing the resolution in the top layer feature maps.

Hinton et al. \cite{Hinton15nips} discovered that the internal structure of the output of a big network can be utilized to distill knowledge from it to a much smaller student network. Through the knowledge distillation between the teacher and the student, the student network learns much better than being trained with the original class labels. We used a similar idea in this paper with a few differences. First, we did not use the output of the big network but the internal feature representation of $conv5$ to guide the learning of the small network. Second, we did not redesign the network architecture of the student network. What we did is to directly drop the first two pooling layer and adjust the stride of the convolution layers. By doing this, we are able to lower the size of the input image to 107x107 while keeping the resolution of the $conv5$ layer intact. The key insight here is that the eliminating pooling and adjusting the stride for convolution does not change the weights inside the network at all. We can directly use the previous weights to initialize the student network. Because the resolution of the $conv5$ layer for both network is the same, given an image the teacher network can automatically generate labels for the student network. Therefore, this approach does not need any label of the input images. We performed the learning with images in the VOC dataset. However, it shall work with any other unlabeled image dataset.

The accelerated network was tested throughout our experiments and we found this approach works very well. Comparing to the bigger network, the performance of the student network does not drop and it is able to achieve state-of-the-art results.

\paragraph{Discussion} The intuition of learning an accelerated tracking network using the internal representation of the teacher network can be thought of as the following metaphor. The teacher network learns rich feature representation of objects by looking at the objects closely (high resolution). Then the teacher network transfers such knowledge to a student at a distance (low resolution) by illustrating different samples of the rich feature representation.

\subsection{Model Details}
We now report other details of the model. The network structure after the $conv5$ layer is simple. One 3x3 convolution layer is used to generate 256 feature maps. Then it is followed by a ReLU layer and two sibling 1x1 convolution layer for the two classification loss defined in  eq. (\ref{eq:track_loss}). Layers after $conv5$ is trained during tracking. The initial 500 positive samples and 5000 negative samples from the first frame was used to train the network for 30 SGD iterations before the tracking starts. For long-term and short-term online model update and other parameters we follow the scheme in \cite{nam2016mdnet}.

\begin{figure}
  \centering
  \subfigure[Precision plots of OTB 50]{
    \includegraphics[scale=0.18]{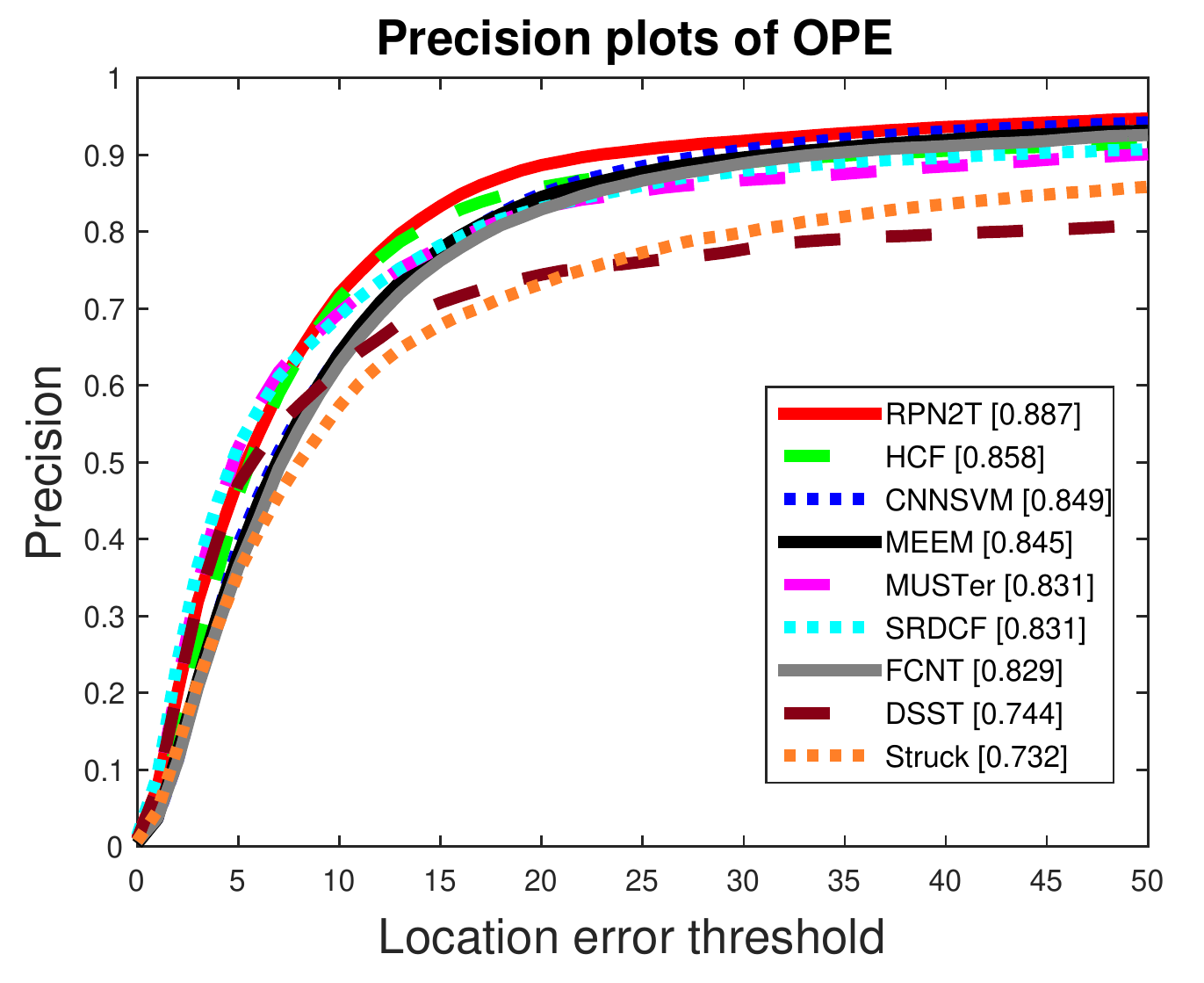}}
  \subfigure[Success plots of OTB 50]{
    \includegraphics[scale=0.18]{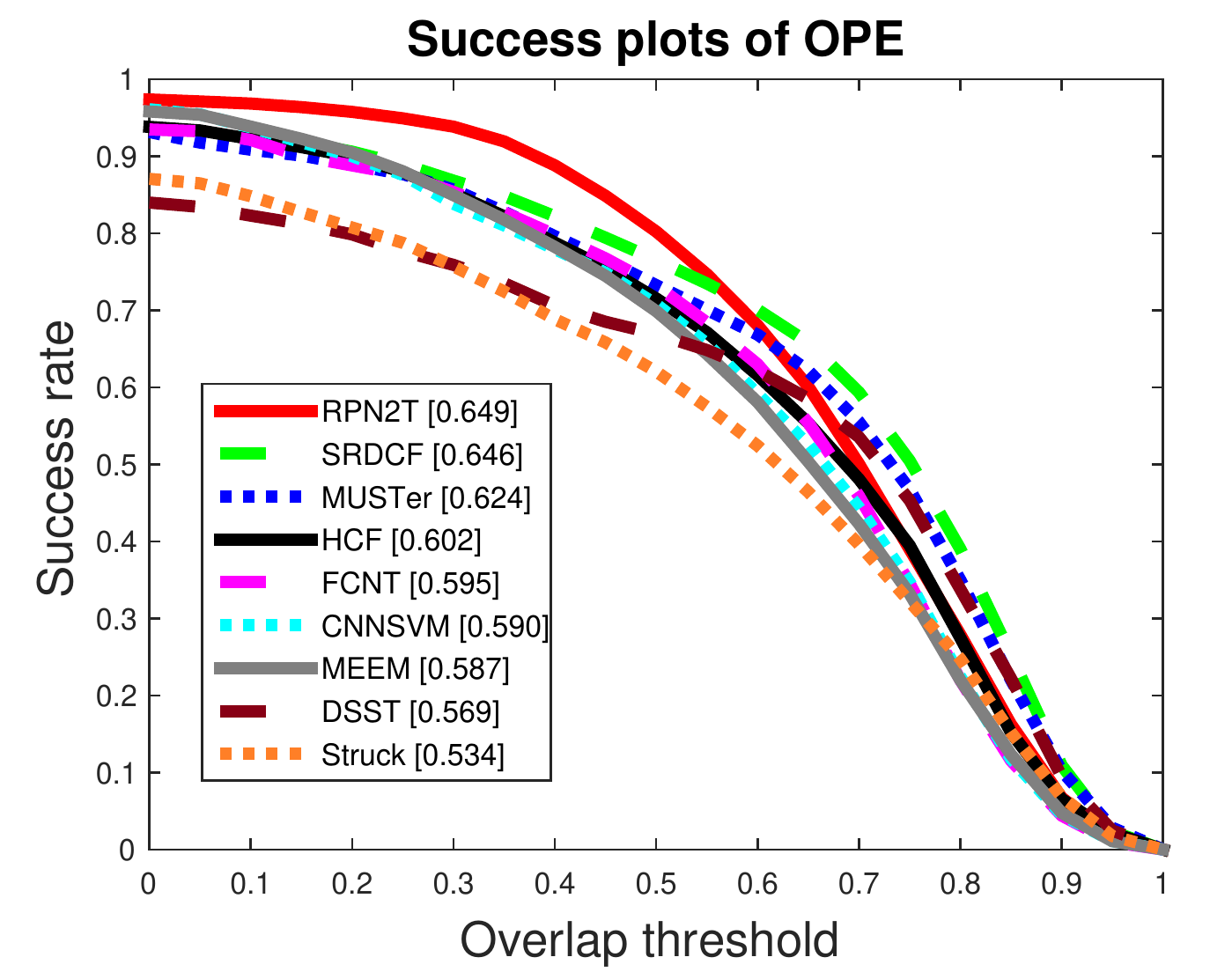}}
  \subfigure[Precision plots of OTB 100]{
    \includegraphics[scale=0.18]{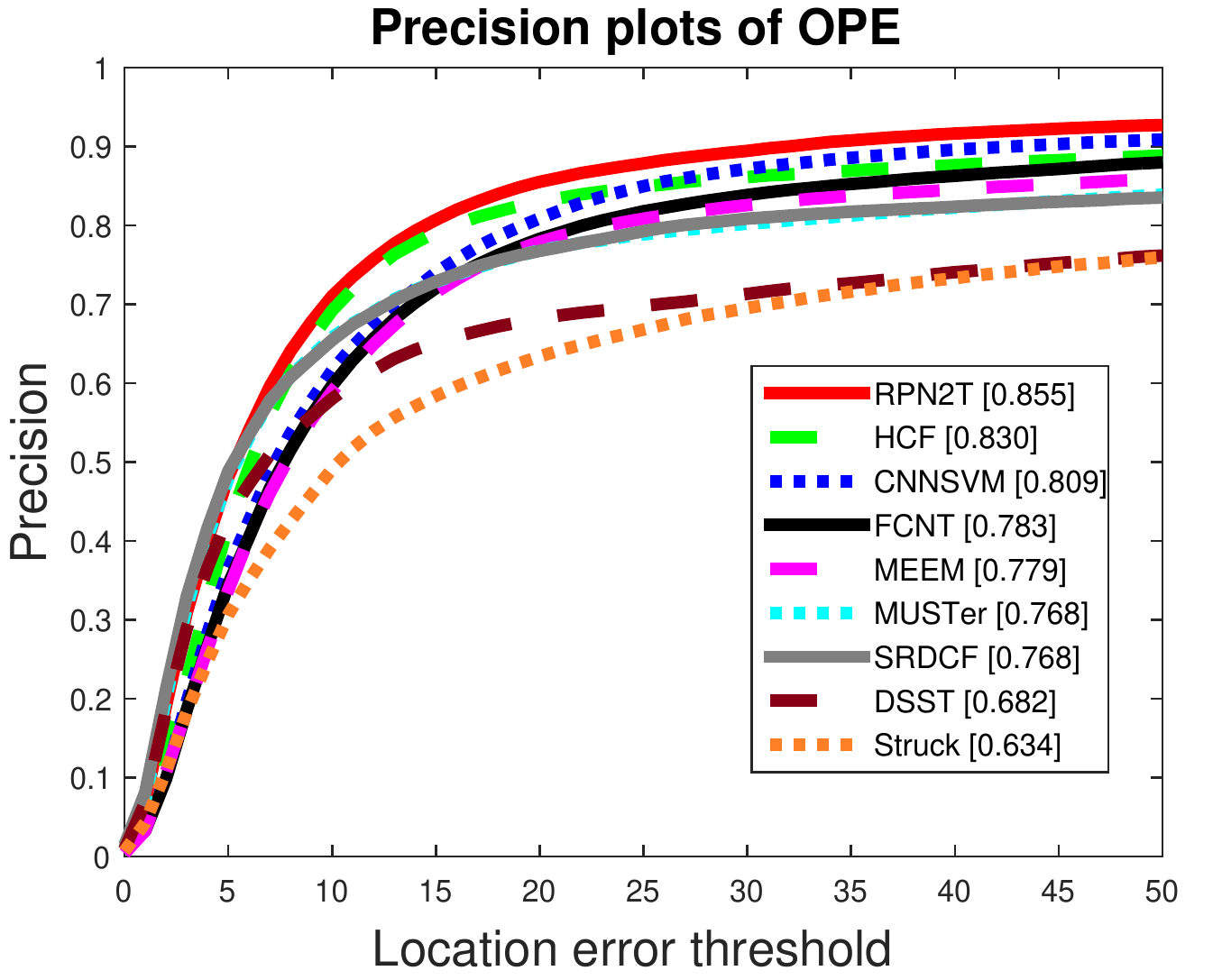}}
		\subfigure[Success plots of OTB 100]{
    \includegraphics[scale=0.18]{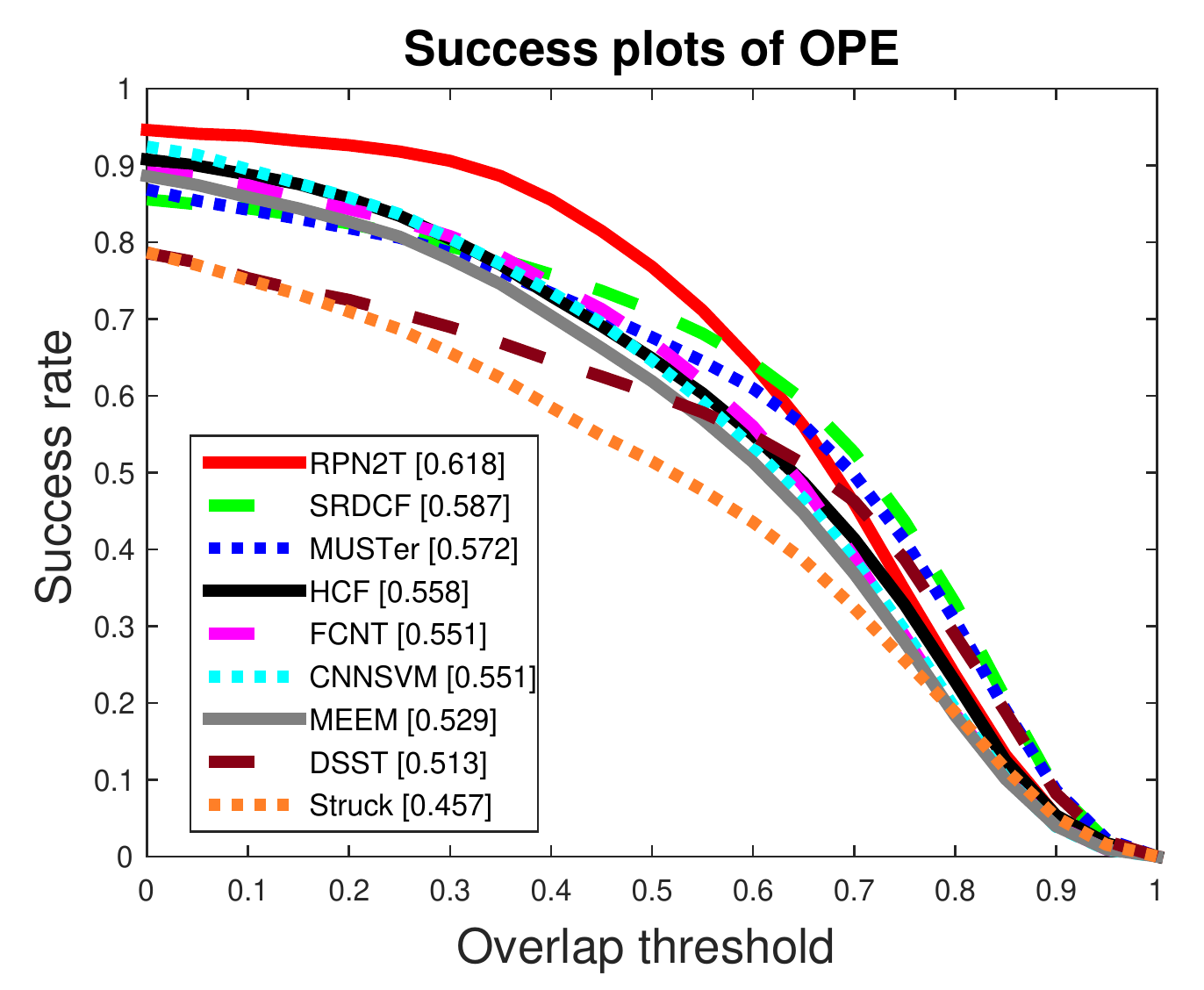}}
\caption{Overall results on OTB 50 and OTB 100.}
\label{fig:otb_results1}
\end{figure}

\section{Experiments}

\begin{figure*}
  \includegraphics[width=1.0\linewidth]{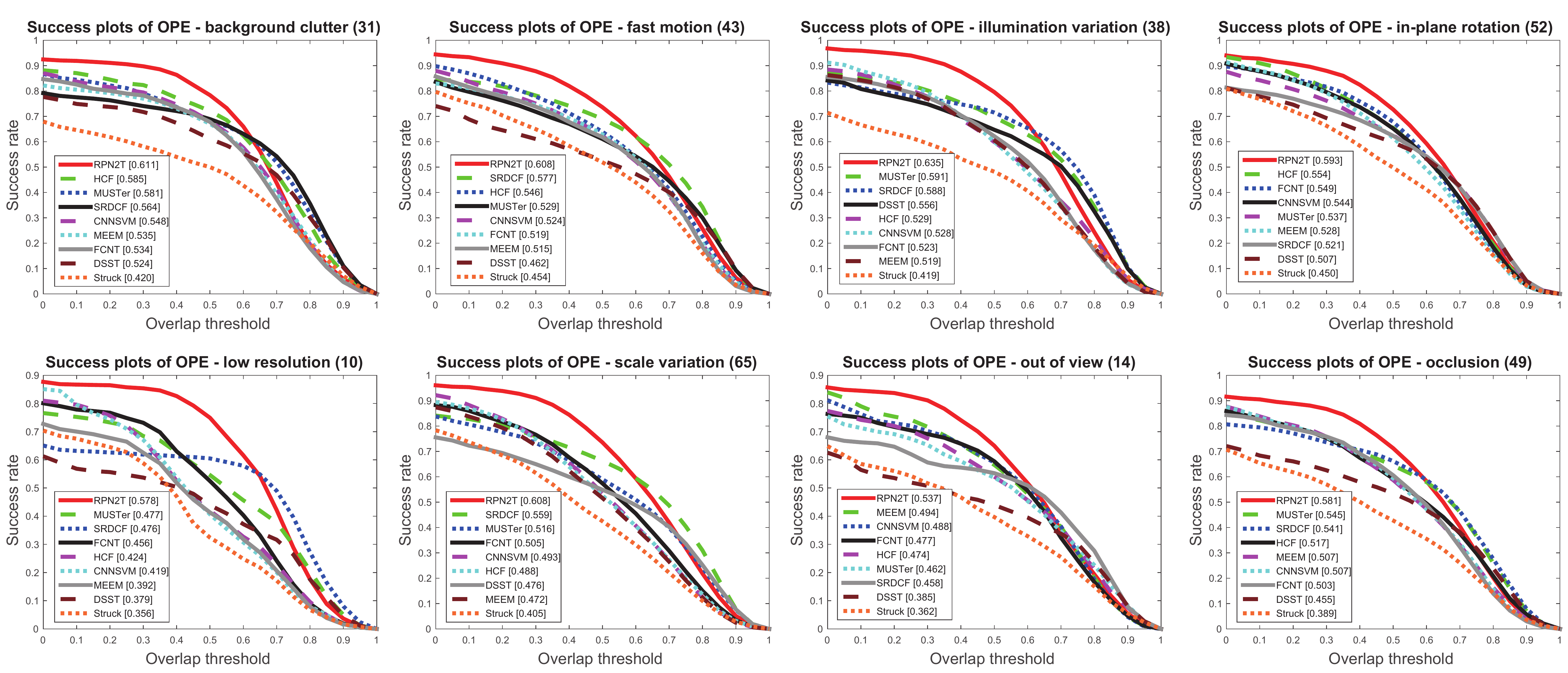}
  \caption{Performance comparison within 8 different challenging scenarios in OTB 100.}
  \label{fig:otb_results2}
\end{figure*}

Our method was implemented in Caffe and its MATLAB bindings. The code was not optimized for speed and the average speed is about 2 FPS with a NVidia GeForce GTX Titan X GPU and Intel i7 3.5GHz CPU. The experiments were conducted with large scale tracking benchmarks including OTB \cite{Wu15pami} and VOT2016 \cite{vot2016results}. A single model with a unified set of parameters was used  in the experiments for all these benchmarks.

\subsection{Results on OTB}
Following the evaluation protocol provided by the OTB benchmark, the performance of trackers was evaluated based on two criterias, namely bounding box overlap ratio and center location error. The final ranks of trackers were determined by the accuracy at the threshold of 20 pixels in the precision plot and the Area Under Curve (AUC) score in the success plot.

Our method was compared with HCF \cite{Ma2015iccv}, CNN-SVM \cite{Hong2015icml}, FCNT \cite{Wang2015iccv}, MEEM \cite{Zhang15eccv}, MUSTer \cite{Hong15cvpr}, SRDCF \cite{Danelljan15iccv}, DSST \cite{Danelljan14bmvc} and Struck \cite{Hare2011iccv}. Among these methods HCF, CNN-SVM and FCNT are CNN based methods, the rest used hand-crafted features. The comparison was carried out for both OTB 50 and OTB 100. MDNet was not included in the comparison because the usage of labeled tracking data in offline training is prohibited in recent standard benchmarks \cite{vot2016}. The overall performance comparison is showed in fig.\ref{fig:otb_results1}. Our method is denoted RPN2T.

We can see from the results that our approach not only outperformed all the methods use hand-crafted features, it also outperformed other CNN based methods. It is worth noting that the size of the network used in our approach is much smaller than the VGG network used in HCF and FCNT. The performance comparison over 8 different challenging scenarios in OTB 100 fig.\ref{fig:otb_results2}. It can be seen that the CNN based methods generally handle the challenging scenarios better than the others. Our method is able to robustly handled all the 8 types of scenarios and outperformed all the other methods.

\begin{figure*}
  \includegraphics[width=1.05\linewidth]{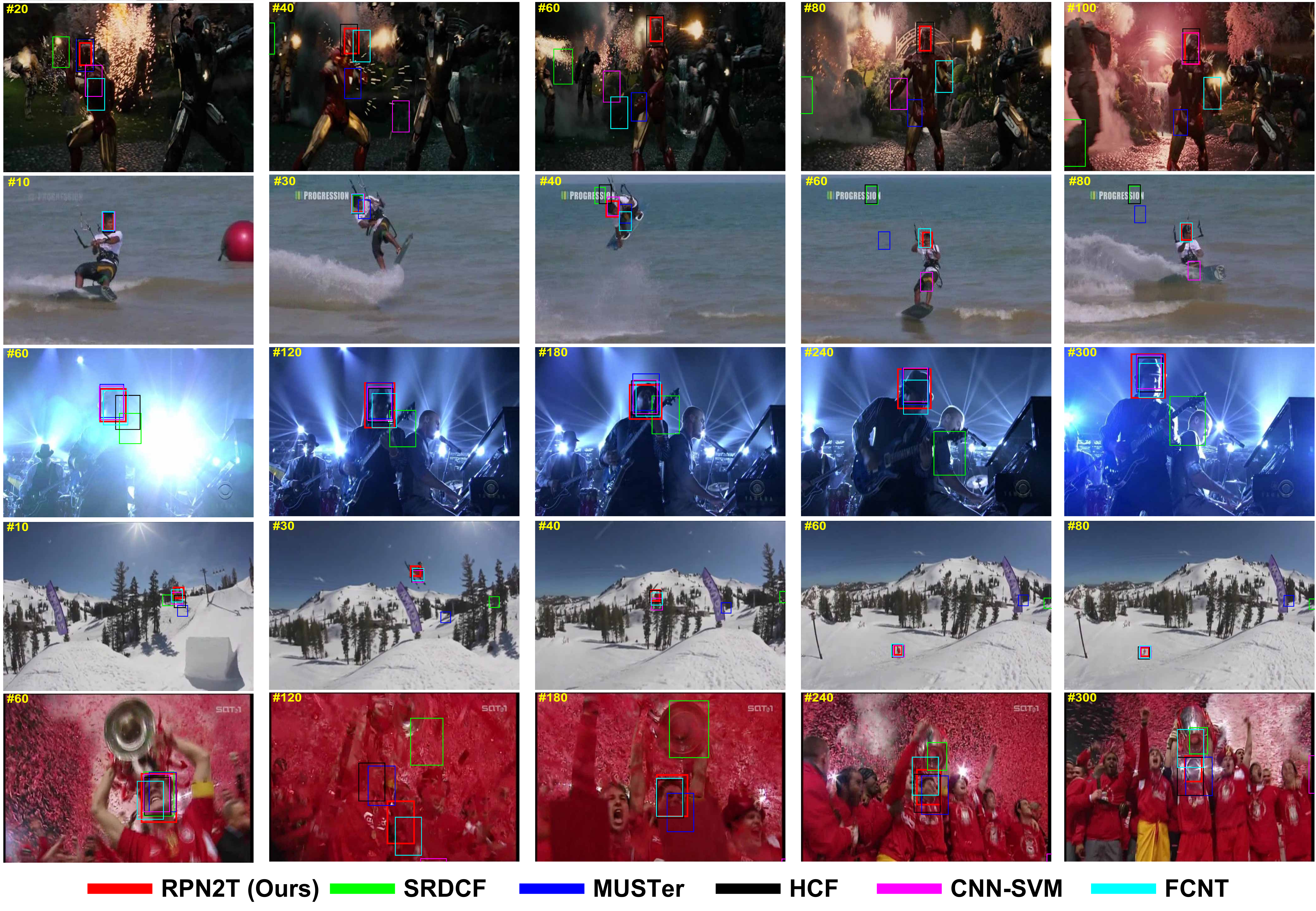}
  \caption{Qualitative results of our method.}
  \label{fig:show_results}
\end{figure*}

Some of the qualitative results can be found in fig.\ref{fig:show_results}.

\subsection{Results on VOT 2016}

\begin{table}[t] 
\caption{Quantitative results on VOT 2016. Comparison to the top 5 methods}
\label{table:vot2016table}
\begin{center}
\begin{small}
\begin{sc}
\begin{tabular}{lccccc}
\toprule
\multirow{1}{*}{Trackers} & \multicolumn{1}{c}{EAO} & \multicolumn{1}{c}{ACC} & \multicolumn{1}{c}{ROB}\\ 
\midrule
C-COT \cite{Danelljan16eccv} &0.331&0.54&0.85\\
TCNN \cite{nam2016tcnn} &0.325&0.55&0.97\\
SSAT &0.321&0.58&1.04\\
MLDF &0.311&0.49&0.83\\
Staple \cite{Bertinetto2016staple} &0.295&0.54&1.35\\
Ours &0.335&0.54&0.87\\
\bottomrule
\end{tabular}
\end{sc}
\end{small}
\end{center}
\vskip -0.1in
\end{table}

Our method was also quantitatively compared with the top 5 methods in the VOT 2016 \cite{vot2016} challange including TCNN \cite{nam2016tcnn}, C-COT \cite{Danelljan16eccv}, Staple \cite{Bertinetto2016staple}, SSAT and MLDF. The VOT 2016 challenge is composed of 60 challenging video sequences with large variations and covers many different scenarios. Differ from the OTB benchmark, the VOT benchmark provides a re-initialization mechanism. Once the trackers fail to track the target, it will be reset with ground-truths in the middle of evaluation process.

In VOT 2016 three primary measures are used to analyze tracking performance namely, accuracy (ACC), robustness (ROB) and expected average overlap (EAO). The accuracy is the average overlap between the predicted and ground truth bounding boxes during successful tracking periods. On the other hand, the robustness measures how many times the tracker loses the target (fails) during tracking. Averaging per-frame accuracy gives per-sequence accuracy, while per-sequence robustness is computed by averaging failure rates
over different runs. The third primary measure, called the expected average overlap (EAO), is an estimator of the average overlap a tracker is expected to attain on a large collection of short-term sequences with the same visual properties as the given dataset. This measure addresses the problem of increased variance and bias of AO \cite{WuLimYang13} measure due to variable sequence lengths on practical datasets. The result is showed in table.\ref{table:vot2016table}.
Our method ranks the first in EAO and is competitive in ACC and ROB.

\section{Conclusion}
In this paper, we showed that by utilizing the internal structure of the top layer feature maps of RPN our method achieved state-of-the-art results in two large scale visual tracking benchmarks without doing ensemble and feature engineering. The key to realize this is a novel loss function which simultaneously considers tracking accuracy and bounding box quality without which the potential of RPN in tracking can not be unleashed.

%
\newpage

\small
\bibliographystyle{plainnat}
\bibliography{egbib}
\end{document}